\documentclass[article,12pt]{article}

\usepackage[affil-it]{authblk}
\usepackage{amsmath}
\usepackage{times}
\usepackage{bm}
\usepackage{bbm}

\usepackage{times}
\usepackage{setspace}
\usepackage{graphicx} 
\usepackage{subfigure} 
\usepackage{algorithm}
\usepackage{algorithmic}

\usepackage{booktabs}
\usepackage[left=2cm,right=2cm,top=2cm,bottom=2cm]{geometry}

\newcommand{\vect}[1]{{\boldsymbol{\mathbf{#1}}}} 
\newcommand{\bfx}{{\vect x}}
\newcommand{\bfv}{{\vect v}}

\newcommand{\bfz}{{\vect z}}

\newcommand{\bfm}{{\vect m}}
\newcommand{\bfzero}{\mathbf{0}}
\newcommand{\bfy}{\mathbf{y}}
\newcommand{\bfmu}{{\vect \mu}}
\newcommand{\bfw}{\mathbf{w}}

\newcommand{\dataset}{\mathcal{D}}

\newcommand{\T}{\top}

\begin{document} 



\title{Local Expectation Gradients for Doubly Stochastic Variational Inference}

\author{Michalis K. Titsias \\ 
Athens University of Economics and Business, \\ 
76, Patission Str. GR10434, Athens, Greece}

\date{}

\maketitle

\begin{abstract} 
We introduce {\em local expectation gradients} which is a general purpose stochastic variational 
inference algorithm for constructing stochastic gradients through sampling from the variational distribution. 
This algorithm divides the problem of estimating the stochastic gradients over multiple variational parameters into smaller sub-tasks 
so that each sub-task exploits intelligently the information coming from the most relevant part 
of the variational distribution. This is achieved by performing an exact expectation over 
the single random variable that mostly correlates with the variational parameter of interest resulting in
a Rao-Blackwellized estimate that has low variance and can work efficiently for both continuous 
and discrete random variables. Furthermore, the proposed algorithm has interesting similarities with Gibbs 
sampling but at the same time, unlike Gibbs sampling, it can be trivially parallelized.     
\end{abstract} 

\section{Introduction}

Stochastic variational inference has emerged as a promising and flexible framework for performing 
large scale approximate inference in complex probabilistic models. It significantly  
extends the traditional variational inference framework \cite{Jordan:1999,bishop:2006:PRML}
by incorporating stochastic approximation \cite{robbinsmonro51} into the optimization of the variational lower bound. 
Currently, there exists two major research directions in stochastic variational inference. The 
first attempts to deal with massive datasets by constructing stochastic gradients 
by using mini-batches of training examples \cite{HoffmanBB10,Hoffmanetal13}. The second direction aims at dealing with the intractable 
expectations under the variational distribution that are encountered in non-conjugate probabilistic models 
\cite{paisleyetal12, Ranganath14, MnihGregor2014, salimans2013, KingmaW13, stochbackpropDeepmind2014, titsiaslazaro2014}. 
The unified idea in the second direction is that stochastic optimization can be carried out by sampling from the variational distribution. This results 
in a {\em doubly stochastic} estimation approach, where the mini-batch source of stochasticity (the first direction above) 
can be combined with a second source of stochasticity associated with sampling from the variational distribution.  
Furthermore, within the second direction there exist now two main sub-classes of methods: the first based on the log derivative trick 
\cite{paisleyetal12, Ranganath14, MnihGregor2014} and the second based on the reparametrization trick
 \cite{KingmaW13, stochbackpropDeepmind2014,titsiaslazaro2014}. The first approach is completely general 
and can allow to apply stochastic optimization of variational lower bounds corresponding to arbitrary models and 
forms for the variational distribution. For instance, probabilistic models having 
both continuous and discrete latent variables can be accommodated by the log derivative trick approach. 
On the other hand, the reparametrization approach is specialised to 
continuous spaces and probabilistic models with differentiable joint distributions. 

In this paper, we are interested to further investigate the doubly stochastic methods that sample from the variational 
distribution. A challenging issue here is concerned with the variance reduction of the stochastic gradients.
Specifically, while the method based on the log derivative trick is the most general one, 
it has been observed to severely suffer from high variance problems \cite{paisleyetal12, Ranganath14, MnihGregor2014}
and thus it is only applicable together with sophisticated variance reduction techniques based on control 
variates. However, the construction of efficient control variates 
is a very challenging issue and in each application depends on the form of the probabilistic model. 
Therefore, it would be highly desirable to investigate whether it is possible to avoid using control variates altogether 
and construct simple stochastic gradient estimates that can work well for any probabilistic model. Notice, that 
while the reparametrization approach \cite{KingmaW13, stochbackpropDeepmind2014,titsiaslazaro2014} 
has shown to perform well without the use of control variates, it is at the same time applicable only
to a restricted class of variational inference problems that involve probabilistic 
models with differentiable joint distributions 
and variational distributions typically taken from the location-scale family.  

Next, we introduce a general purpose algorithm for constructing stochastic gradients
by sampling from the variational distribution that has a very low variance and can work efficiently
without the need of any variance reduction technique. This method builds upon the 
log derivative trick and it is based on the key observation that stochastic gradient estimation
over multiple variational parameters can be divided into smaller sub-tasks 
where each sub-task requires using more information coming from some part 
of the variational distribution and less information coming from other parts. For instance, assume we have a factorized variational distribution 
of the form $\prod_{i=1}^n q_{v_i}(x_i)$ where $v_i$ is a local variational 
parameter and $x_i$ the associated latent variable. Clearly,  $v_i$ determines the distribution over $x_i$,
and therefore we expect the latent variable $x_i$ to be the most important piece 
of information for estimating $v_i$ or its gradient. Based on this intuitive observation
we introduce the {\em local expectation gradients} algorithm that provides a stochastic gradient over $v_i$ by performing an exact expectation over 
the associated random variable $x_i$ while using a single sample from the remaining 
latent variables. Essentially this consist of a Rao-Blackwellized estimate that allows to dramatically 
reduce the variance of the stochastic gradient so that, for instance, for continuous spaces the new stochastic gradient
is guaranteed to have lower variance than the stochastic gradient corresponding to the 
state of the art reparametrization method. Furthermore, the local expectation algorithm has striking 
similarities with Gibbs sampling with the important difference, that unlike Gibbs sampling,  it can be trivially parallelized.     

The remainder of the paper has as follows: Section \ref{eq:svi} discusses the main two  
types of algorithms for stochastic variational inference that are based on simulating from the variational
distribution. Section \ref{sec:localexp} describes the proposed local expectation gradients algorithm. 
Section \ref{sec:experiments} provides experimental results and the paper concludes 
with a discussion in Section \ref{sec:discussion}.

\section{Stochastic variational inference \label{eq:svi}}

Here, we discuss the main ideas behind current algorithms on stochastic variational inference 
and particularly doubly stochastic methods that sample from the variational distribution in order to approximate 
intractable expectations using Monte Carlo. Given a joint probability distribution $p(\bfy,\bfx)$ where 
$\bfy$ are observations and $\bfx$ are latent variables (possibly including model parameters that consist of 
random variables) and a variational distribution $q_{\bfv}(\bfx)$, the objective is to maximize
the lower bound 
\begin{align}
F(\bfv) & = \mathbbm{E}_{q_{\bfv}(\bfx)} \left[
\log p(\bfy,\bfx)  - \log q_{\bfv}(\bfx) \right],  \label{eq:bound1} \\
& = \mathbbm{E}_{q_{\bfv}(\bfx)} \left[
\log p(\bfy,\bfx) \right]  - \mathbbm{E}_{q_{\bfv}(\bfx)} \left[ \log q_{\bfv}(\bfx) \right],  
\label{eq:bound2}
\end{align}
with respect to the variational parameters $\bfv$. Ideally, in order to tune 
$\bfv$ we would like to have a closed-form expression for the lower bound so that we could  
subsequently maximize it by using standard optimization routines such as gradient-based algorithms.  
However, for many probabilistic models and forms of the variational 
distribution at least one of the two expectations in (\ref{eq:bound2}) are intractable. 
For instance, if $p(\bfy,\bfx)$ is defined though a neural network, the expectation
 $\mathbbm{E}_{q_{\bfv}(\bfx)} \left[\log p(\bfy,\bfx) \right]$ will  be analytically intractable 
despite the fact that $q_{\bfv}(\bfx)$ might have a very simple form, 
such as a Gaussian, so that the second expectation in eq.\ (\ref{eq:bound2}) (i.e.\ the entropy of $q_{\bfv}(\bfv)$) 
will be tractable. In other cases, such as when $q_{\bfv}(\bfv)$ is a mixture model or is defined through a
complex graphical model, the entropic term will also be intractable. Therefore, in general we are 
facing with the following intractable expectation 
\begin{equation}
\widetilde{F}(\bfv) = \mathbbm{E}_{q_{\bfv}(\bfx)} \left[ f(\bfx) \right],
\label{eq:genexpect}
\end{equation}
where $f(\bfx)$ can be either $\log p(\bfy,\bfx)$, $-\log q_{\bfv}(\bfx)$  or 
$\log p(\bfy,\bfx)  - \log q_{\bfv}(\bfx)$, from which we would like to efficiently 
estimate the gradient over $\bfv$ in order to apply gradient-based optimization. 

The most general method for estimating the gradient $\nabla_{\bfv} \widetilde{F}(\bfv)$
is based on the log derivative trick \cite{paisleyetal12, Ranganath14, MnihGregor2014}.
Specifically, this makes use of the property $\nabla_{\bfv} q_{\bfv}(\bfx) = q_{\bfv}(\bfx) \nabla_{\bfv} \log q_{\bfv}(\bfx)$, 
which allows to write the gradient as  
\begin{equation}
\nabla_{\bfv} \widetilde{F}(\bfv) = \mathbbm{E}_{q_{\bfv}(\bfx)} \left[ f(\bfx) \nabla_{\bfv} \log q_{\bfv}(\bfx) \right]
\label{eq:gengrad}
\end{equation}
and then obtain an unbiased estimate according to 
\begin{equation}
\frac{1}{S} \sum_{s=1}^S f(\bfx^{(s)}) \nabla_{\bfv} \log q_{\bfv}(\bfx^{(s)}),
\label{eq:logderivestimate}
\end{equation}
where each $\bfx^{(s)}$ is an independent draw from $q_{\bfv}(\bfx)$. While 
this estimate is unbiased, it has been observed to severely suffer from high variance so that  
in practice it is necessary to consider variance reduction techniques such as those based on 
control variates \cite{paisleyetal12, Ranganath14, MnihGregor2014}. Despite this limitation the above framework is 
very general as it can deal with any variational distribution over both discrete and
continuous latent variables. The proposed method presented in Section \ref{sec:localexp} 
is essentially based on the log derivative trick, but it does not suffer from high variance. 

The second approach is suitable for continuous spaces where $f(\bfx)$ is a differentiable function
of $\bfx$ \cite{KingmaW13, stochbackpropDeepmind2014,titsiaslazaro2014}. It is based on a simple transformation 
of (\ref{eq:genexpect}) which allows to move the variational parameters $\bfv$ inside $f(\bfx)$
so that eventually the expectation is taken over a base distribution that does not depend on the 
variational parameters anymore. For example, if the variational 
distribution is the Gaussian $\mathcal{N}(\bfx|\bfmu,L L^{\T})$ where $\bfv = (\bfmu,L)$, the expectation in 
(\ref{eq:genexpect}) can be re-written as $\widetilde{F}(\bfmu,L) = \int \mathcal{N}(\bfz|\bfzero, I) f(\bfmu + L \bfz) d \bfz$
and subsequently the gradient over $(\bfmu,L)$ can be approximated by the following unbiased Monte Carlo estimate 
\begin{equation}
\frac{1}{S} \sum_{s=1}^S \nabla_{(\bfmu,L)} f(\bfmu + L \bfz^{(s)}),
\label{eq:reparamEstimate}
\end{equation}
where each $\bfz^{(s)}$ is an independent sample from $\mathcal{N}(\bfz|\bfzero,I)$.
This estimate makes efficient use of the slope of $f(\bfx)$ which allows to perform informative moves in 
the space of $(\bfmu,L)$. For instance, observe that as $L \rightarrow 0$ the gradient over 
$\bfmu$ approaches $\nabla_{\bfmu} f(\bfmu)$ so that the optimization reduces to a standard gradient-ascent 
procedure for locating a mode of $f(\bfx)$. Furthermore, it has been shown experimentally in several studies 
\cite{KingmaW13, stochbackpropDeepmind2014,titsiaslazaro2014} that the estimate in (\ref{eq:reparamEstimate}) 
has relatively low variance and can lead to efficient optimization even when a single sample is 
used at each iteration. Nevertheless, a limitation of the approach is that it is only applicable to models where $\bfx$ is continuous 
and $f(\bfx)$ is differentiable. Even within this subset of models we are also additionally 
restricted to using certain classes of variational distributions \cite{KingmaW13, stochbackpropDeepmind2014}.
    
Therefore, it is clear that from the current literature is lacking a universal method that both can be applicable to a very broad class of models 
(in both discrete and continuous spaces) and also provide low-variance stochastic gradients. 
Next, we introduce such an approach..

\section{Local expectation gradients \label{sec:localexp}} 

Suppose that the $n$-dimensional latent vector $\bfx$ in the probabilistic model takes values in some space $\mathcal{S}_1 \times \ldots \mathcal{S}_n$ 
where each set $\mathcal{S}_i$ can be continuous or discrete. We consider a variational distribution over $\bfx$ 
that is represented as a directed graphical model having the following joint density   
\begin{equation}
q_{\bfv}(\bfx) = \prod_{i=1}^n q_{v_i}(x_i|\text{pa}_i),
\label{eq:vardist} 
\end{equation}
where $q_{v_i}(x_i|\text{pa}_i)$ is the conditional factor over $x_i$ given the set of the parents denoted by $\text{pa}_i$. 
We assume that each conditional factor has its own separate set of variational parameters $v_i$ and $\bfv = (v_i,\ldots,v_n)$. 
The objective is then to obtain a stochastic approximation for the gradient of the lower bound over each variational 
parameter $v_i$ based on the log derivative form in eq.\ (\ref{eq:gengrad}).  

Our method is motivated by the observation that each parameter $v_i$ 
is rather influenced mostly by its corresponding latent variable $x_i$ since $v_i$  
determines the factor $q_{v_i}(x_i|\text{pa}_i)$. 
Therefore, to get information about the gradient of 
$v_i$ we should be exploring multiple possible values of $x_i$
and a rather smaller set of values from the remaining latent variables 
$\bfx_{\setminus i}$. Next we take this idea into the extreme 
where we will be using infinite draws from $x_i$ (i.e.\ essentially an exact expectation) 
together with just a single sample of $\bfx_{\setminus i}$. More precisely, we factorize the 
variational distribution as follows
\begin{equation}
q_{\bfv}(\bfx) = q(x_i|\text{mb}_i) q(\bfx_{\setminus i}), 
\end{equation}
where $\text{mb}_i$ denotes the Markov blanket of $x_i$. By using the log derivative trick 
the gradient over $v_i$ can be written as 
\begin{align}
\nabla_{v_i} \widetilde{F}(\bfv) & = 
\mathbbm{E}_{q (\bfx)} \left[ f(\bfx) \nabla_{v_i} \log q_{v_i}(x_i|\text{pa}_i) \right], \nonumber \\
& = \mathbbm{E}_{q (\bfx_{\setminus i})} \left[ \mathbbm{E}_{q(x_i|\text{mb}_i)} \left[ f(\bfx) \nabla_{v_i} \log q_{v_i}(x_i|\text{pa}_i) \right] \right], 
\label{eq:graditer}
\end{align}
where in the second expression we used the law of iterated expectations. Then, an unbiased 
stochastic gradient, say at the $t$-th iteration of an optimization algorithm, can be obtained by drawing a single sample
$\bfx_{\setminus i}^{(t)}$ from $q(\bfx_{\setminus i})$ so that 
\begin{equation}
\mathbbm{E}_{q(x_i|\text{mb}_i^{(t)})} \left[ f(\bfx_{\setminus i}^{(t)}, x_i) \nabla_{v_i} \log q_{v_i}(x_i|\text{pa}_i^{(t)}) \right],  
\label{eq:graditerGibbs}
\end{equation}
which is the expression for the proposed stochastic gradient for the parameter $v_i$.  
To get an independent sample $\bfx_{\setminus i}^{(t)}$  from $q(\bfx_{\setminus i})$ 
we can simply simulate a full latent vector $\bfx^{(t)}$ from $q_{\bfv}(\bfx)$ by applying the standard 
ancestral sampling procedure for directed graphical models \cite{bishop:2006:PRML}. 
Then, the sub-vector $\bfx_{\setminus i}^{(t)}$ is by construction an independent draw from the marginal $q(\bfx_{\setminus i})$. Furthermore, 
the sample $\bfx^{(t)}$ can be thought of as a {\em pivot} sample that is needed to be drawn once 
and then it can be re-used multiple times in order to compute all stochastic 
gradients for all variational parameters $v_1,\ldots,v_n$ according to 
eq.\ (\ref{eq:graditerGibbs}).
 
When the variable $x_i$ takes discrete values, the expectation under 
$q(x_i|\text{mb}_i^{(t)})$ in eq.\ (\ref{eq:graditerGibbs}) reduces to a sum of terms
associated with all possible values of $x_i$.   
On the other hand, when $x_i$ is a continuous variable the expectation in (\ref{eq:graditerGibbs}) corresponds to 
an univariate integral that in general may not be analytically tractable. In this case we shall use fast numerical 
integration methods (e.g.\ Gaussian quadrature when $q_{v_i}(x_i|\text{pa}_i)$ is Gaussian). 

We shall refer to the above algorithm for providing stochastic gradients over variational parameters
as {\em local expectation gradients} and pseudo-code of a stochastic variational inference scheme that internally uses this algorithm 
is given in Algorithm \ref{alg1}. Notice that Algorithm \ref{alg1} corresponds to the case where
$f(\bfx) = \log p(\bfy,\bfx) - \log q_{\bfv}(\bfx)$ while other cases can be expressed similarly.

In the next two sections we discuss the computational complexity of the proposed algorithm and draw interesting connections between local 
expectation gradients with Gibbs sampling (Section \ref{sec:conneGibbs}) and the reparametrization approach for differentiable 
functions $f(\bfx)$ (Section \ref{sec:conneReparam}). 

\begin{algorithm}[tb]
   \caption{Stochastic variational inference using local expectation gradients\label{algor1}}
   \label{alg1}
\begin{algorithmic}
   \STATE {\bfseries Input:} $f(\bfx)$, $q_{\bfv}(\bfx)$.
   \STATE Initialize $\bfv^{(0)}$, $t=0$.
   \REPEAT
   \STATE Set $t=t+1$.
   \STATE Draw pivot sample $\bfx^{(t)} \sim q_{\bfv}(\bfx)$.
   \FOR{$i=1$ {\bfseries to} $n$}
    \STATE $dv_i = \mathbbm{E}_{q(x_i|\text{mb}_i^{(t)})} \left[ f(\bfx_{\setminus i}^{(t)}, x_i) \nabla_{v_i} \log q_{v_i}(x_i|\text{pa}_i^{(t)}) \right]$.
    \STATE $v_i = v_i + \eta_t dv_i$.
   \ENDFOR  
   \UNTIL{convergence criterion is met.}
\end{algorithmic}
\end{algorithm}

\subsection{Time complexity and connection with Gibbs sampling \label{sec:conneGibbs}}

In this section we discuss computational issues when running the proposed algorithm
and we point out similarities and difference that has with Gibbs sampling. 

Let us assume that time complexity is dominated by function evaluations of 
$f(\bfx)$. We further assume that this function neither factorizes into a sum of local terms,
where each term depends on a subset of variables, nor allows savings 
by performing incremental updates of statistics computed during intermediate 
steps when evaluating $f(\bfx)$. Based on this the complexity per iteration is $O(n K)$ where $K$ is the maximum number 
of evaluations of $f(\bfx)$ needed when estimating the gradient for each $v_i$.
If each $x_i$ takes $K$ discrete values, the exact number of 
evaluations will be $n (K-1) + 1$  where the ``plus one'' comes from the evaluation of 
the pivot sample $\bfx^{(t)}$ that needs to be performed once and then it can be re-used for 
each $i=1,\ldots,n$. Notice that this time complexity corresponds to a  worst-case scenario. In 
practice, we could save a lot of computations by taking advantage of
any factorization in $f(\bfx)$ and also the fact that any function evaluation 
is performed for inputs that are the same as the pivot sample $\bfx^{(t)}$ 
but having a single variable changed. Furthermore, once we have drawn the pivot 
sample $\bfx^{(t)}$ all function evaluations can be trivially parallelized.

There is an interesting connection between local expectation gradients and  
Gibbs sampling. In particular, carrying out Gibbs sampling in the variational distribution in 
eq.\ (\ref{eq:vardist}) requires iteratively sampling from each conditional 
$q(x_i|\text{mb}_i)$, for $i=1,\ldots,n$. Clearly, the same conditional  
appears also in the local expectation algorithm when estimating the stochastic gradients. 
The obvious difference is that instead of sampling from  $q(x_i|\text{mb}_i)$ 
we now average under this distribution. Furthermore, for models having discrete 
latent variables the time complexity per iteration is the same as with Gibbs 
sampling with the important difference, however, that the algorithm of local expectation gradients
is trivially parallelizable while Gibbs sampling is not.   
 
\subsection{Connection with the reparametrization approach \label{sec:conneReparam}}

A very interesting property of local expectation gradients is that it is guaranteed to provide stochastic gradients 
having lower variance that the state of the art reparametrization method \cite{KingmaW13, stochbackpropDeepmind2014,titsiaslazaro2014}, 
also called stochastic belief propagation \cite{stochbackpropDeepmind2014}, which is suitable for continuous 
spaces and differential functions $f(\bfx)$. Specifically, we will prove
that this property holds for any factorized location-scale variational distribution of the form 
\begin{equation}
q_{\bfv}(\bfx) =  \prod_{i=1}^n q_{v_i} (x_i),
\end{equation}
where $v_i = (\mu_i, \ell_i)$, $\mu_i$ is a location-mean parameter and $\ell_i$ is a scale parameter. 
Given that $q(z_i)$ is the base distribution based on which we can reparametrize $x_i$ according to 
$x_i = \mu_i + \ell z_i$ with $z_i \sim q(z_i)$, the single-sample stochastic gradient over $v_i$
is given by 
\begin{equation}
\nabla_{v_i} f(\bfmu + \boldsymbol{\ell} \circ \bfz^{(t)}), \ \ z^{(t)}_i \sim q(z_i),
\label{eq:gradReparFactor} 
\end{equation}
where $\bfmu$ is the vector of all $\mu_i$s, similarly $\boldsymbol{\ell}$ and $\bfz^{(t)}$ are vectors of $\ell_i$s 
and $z_i^{(t)}$s, while $\circ$ denotes element-wise product. The local expectation stochastic gradient 
from eq.\ (\ref{eq:graditerGibbs}) takes the form 
\begin{align}
& \int q_{v_i} (x_i) f(\bfx_{\setminus i}^{(t)}, x_i) \nabla_{v_i} \log q_{v_i}(x_i) d x_i, \nonumber \\
& \int \nabla_{v_i} q_{v_i} (x_i) f(\bfx_{\setminus i}^{(t)}, x_i) d x_i. 
\label{eq:gradLocExpFactor1}
\end{align}
Now notice that $\bfx_{\setminus i}^{(t)}= \bfmu_{\setminus i} + \boldsymbol{\ell}_{\setminus i} \circ \bfz_{\setminus i}^{(t)}$. Also 
by interchanging the order of the gradient and integral operators together with using the base distribution we have  
\begin{align}
& \nabla_{v_i} \int q_{v_i} (x_i) f(\bfmu_{\setminus i} + \boldsymbol{\ell}_{\setminus i} \circ \bfz_{\setminus i}^{(t)}, x_i) d x_i, \nonumber \\
& = \nabla_{v_i} \int q(z_i) f(\bfmu + \boldsymbol{\ell} \circ \bfz^{(t)}) d z_i,  \nonumber \\
& = \int q(z_i) \nabla_{v_i} f(\bfmu + \boldsymbol{\ell} \circ \bfz^{(t)}) d z_i,
\label{eq:gradLocExpFactor2}
\end{align}
where the final eq.\ (\ref{eq:gradLocExpFactor2}) is clearly an expectation of the reparametrization gradient from 
eq.\ (\ref{eq:gradReparFactor}). Therefore, based on the standard Rao-Blackwellization argument the 
variance of the stochastic gradient obtained by (\ref{eq:gradLocExpFactor2}) will always be lower or equal than the variance of the gradient  
of the reparametrization method. To intuitively understand this, observe that eq.\ (\ref{eq:gradLocExpFactor2})
essentially says that the single-sample reparametrization gradient from (\ref{eq:gradReparFactor}) is just a Monte Carlo approximation 
to the local expectation stochastic gradient obtained by drawing a single sample from the base distribution, thus 
naturally it should have higher variance. In the experiments in Section \ref{sec:experiments} we typically observe 
that the variance provided by local expectation gradients is roughly one order of magnitude lower 
than the corresponding variance of the reparametrization gradients. 


\section{Experiments \label{sec:experiments}} 

In this section, we apply local expectation gradients (LeGrad) to different types of stochastic variational inference 
problems and we compare it against the standard stochastic gradient based on the log derivative trick (LdGrad)
described by eq.\ (\ref{eq:logderivestimate}) as well as the reparametrization-based gradient (ReGrad) given by eq.\ 
(\ref{eq:reparamEstimate}). In Section \ref{sec:expGaussian} we consider fitting using a factorized variational 
Gaussian distribution a highly correlated multivariate Gaussian. In Section \ref{sec:logreg},
we consider a two-class classification problem using two digits from the MNIST database and we approximate 
a Bayesian logistic regression model using stochastic variational inference. Finally, in Section 
\ref{sec:sigmoidbnet} we consider a sigmoid belief network with one layer of hidden variables
and we fit it to the binarized version of the MNIST digits. For this problem we also parametrize 
the variational distribution using a recognition model.    
 
\subsection{Fitting a high dimensional Gaussian  \label{sec:expGaussian}}

\begin{figure}[!htb]
\vskip 0.2in
\begin{center}
{\includegraphics[scale=0.4]  
{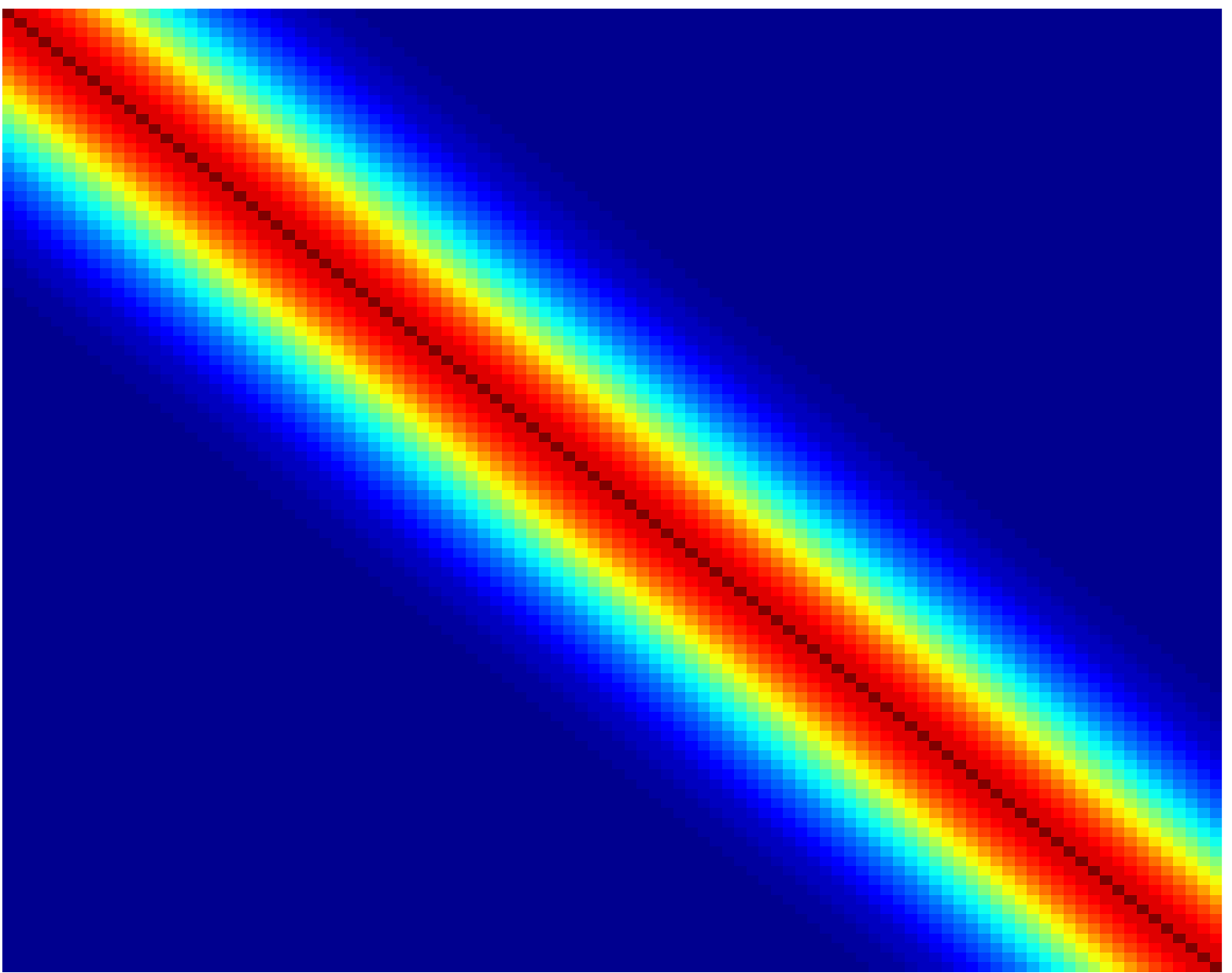}} 
\caption{The $100 \times 100$ covariance matrix $\Sigma$ used in the experiment in Section \ref{sec:expGaussian}.} 
\label{fig:covmatrix}
\end{center}
\vskip -0.2in
\end{figure}

We start with a simple ``artificial'' variational inference problem where we would like to 
fit a factorized variational Gaussian distribution of the form 
\begin{equation}
q_{\bfv}(\bfx) = \prod_{i=1}^n \mathcal{N}(x_i|\mu_i, \ell_i^2),
\label{eq:factorGauss} 
\end{equation}
to a highly correlated multivariate Gaussian of the form $\mathcal{N}(\bfx|\bfm, \Sigma)$. 
We assume that $n=100$, $\bfm = \boldsymbol{2}$, where $\boldsymbol{2}$ is the $100$-dimensional vector 
of $2$s. Further, the correlated matrix $\Sigma$ was constructed from a kernel function 
so that $\Sigma_{ij} = e^{ - \frac{1}{2} (x_i - x_j)^2}  + 0.1 \delta_{i j}$ and
where the inputs where placed in an uniform grid in $[0,10]$. This covariance matrix is shown in Figure 
\ref{fig:covmatrix}. The smaller eigenvalue of $\Sigma$ is roughly $0.1$ so we expect 
that the optimal values for the variances $\ell_i^2$ to be around $0.1$  
(see e.g.\ \cite{bishop:2006:PRML}) while the optimal value for each $\mu_i$ is $2$. 

Given that the latent vector $\bfx$ is continuous, to obtain the stochastic 
gradient for each $(\mu_i,\ell_i)$ we need to apply numerical integration. More precisely,
the stochastic gradient for LeGrad according to eq.\ (\ref{eq:graditerGibbs}) reduces to an expectation 
under the Gaussian $\mathcal{N}(x_i|\mu_i, \ell_i^2)$ and therefore we can naturally 
apply Gaussian quadrature. We used the quadrature rule having $K=5$ grid points\footnote{Gaussian 
quadrature with $K$ grid points integrates exactly polynomials  up to $2 K -1$ degree.} so that the whole 
complexity of LeGrad was $O(n K) = O(500)$ function evaluations per iteration (see Section
\ref{sec:conneGibbs}). When we applied the standard LdGrad approach we set the number 
of samples in (\ref{eq:logderivestimate}) equal to $S=500$ so that the computational 
costs of LeGrad and LdGrad match exactly with one another. When using the ReGrad 
approach based on (\ref{eq:reparamEstimate}) we construct the stochastic 
gradient using a single sample, as this is typical among the practitioners that use this method, 
and also because we want to empirically confirm the theory 
from Section \ref{sec:conneReparam} which states that the LeGrad method 
should always have lower variance than ReGrad given that the latter uses 
a sample size of one. 

\begin{figure*}[!htb]
\vskip 0.2in
\begin{center}
\begin{tabular}{cccc}
{\includegraphics[scale=0.3]
{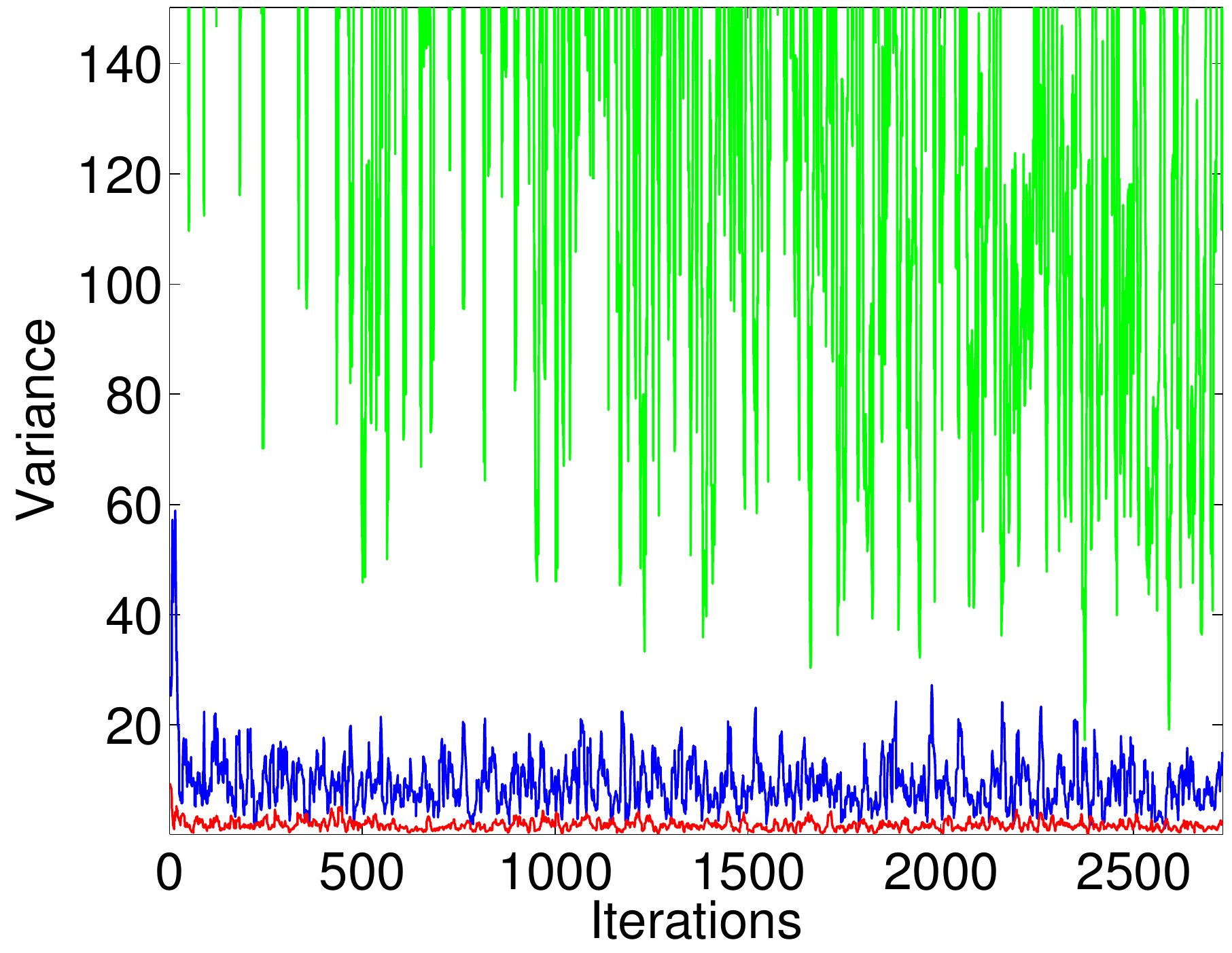}}  &
{\includegraphics[scale=0.3]  
{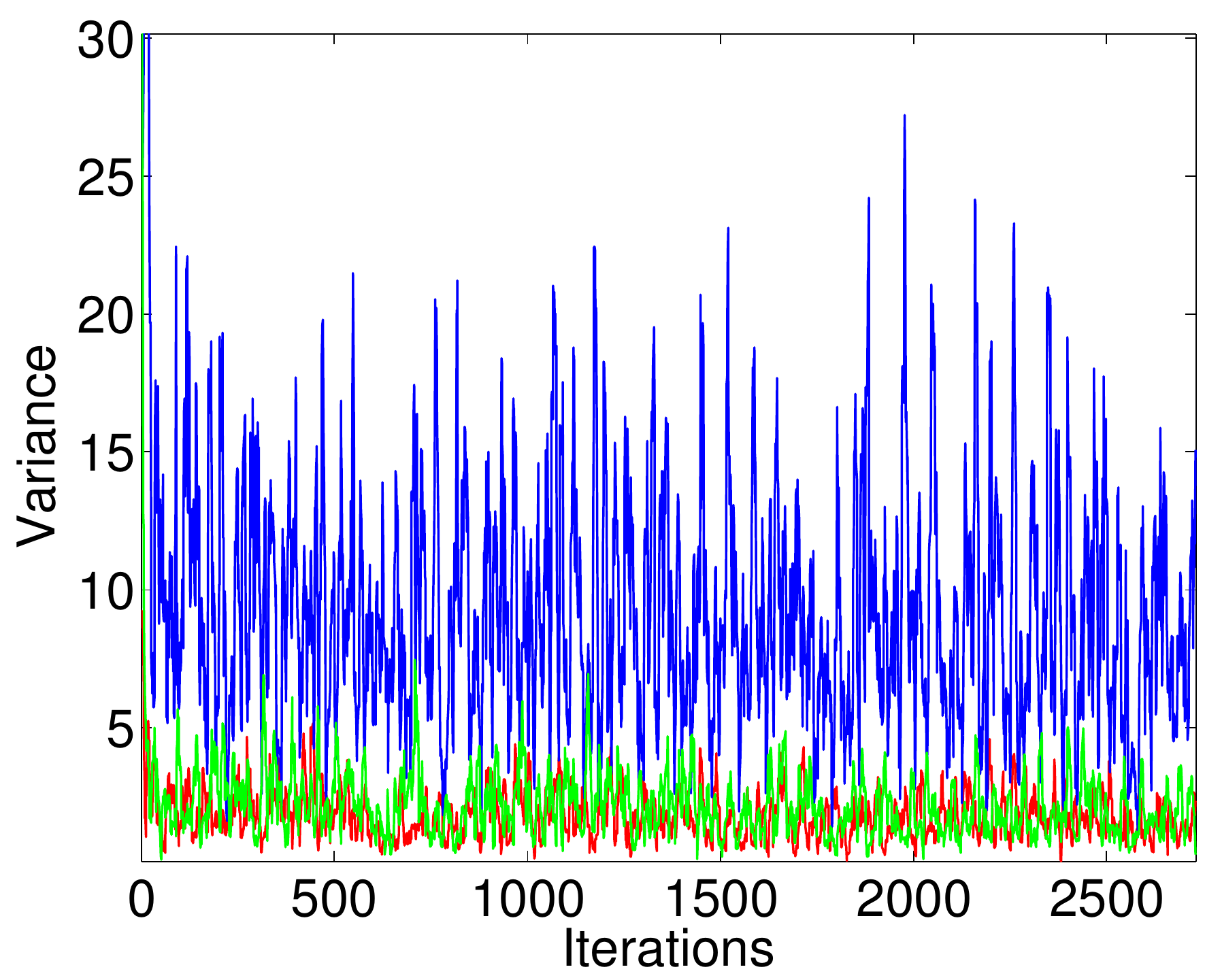}} &
{\includegraphics[scale=0.3]
{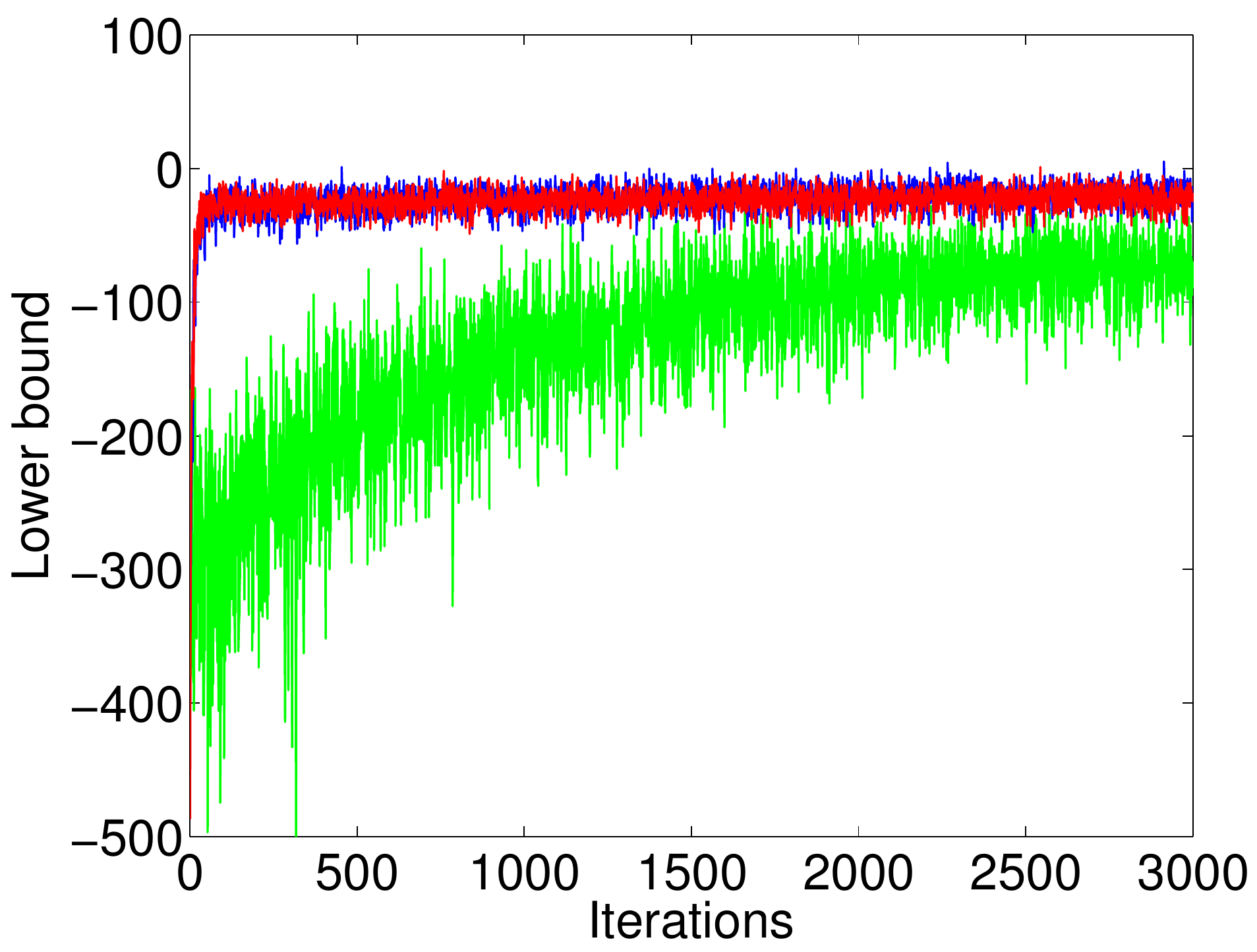}} \\
(a) & (b) & (c)
\end{tabular}
\caption{The panel in (a) shows the variance of the gradient for the variational parameter $\mu_1$ 
when using LeGrad (red line), ReGrad (blue line) and LdGrad (green line). The number of samples 
used by LdGrad was $S=500$. The panel in (b) shows again the variances for all methods (the red and blue lines are from 
(a)) when LdGrad uses $S=10^4$ samples. The panel in (c) shows the evolution of the 
stochastic value of the lower bound (for LdGrad $S=500$ was used).} 
\label{fig:toyGaussian}
\end{center}
\vskip -0.2in
\end{figure*}

Figure \ref{fig:toyGaussian}(a) shows the evolution of the variance of the three alternative stochastic 
gradients (estimated by using the first variational parameter $\mu_1$ and a running window 
of $10$ previous iterations) as the stochastic optimization algorithm iterates. Clearly,  
LeGrad (red line) has the lowest variance, then comes ReGrad (blue line) and last 
is LdGrad which despite the fact that it uses $500$ independent samples in the Monte Carlo average in
eq.\ (\ref{eq:logderivestimate}) suffers from high variance. Someone could 
ask about how many samples the LdGrad method requires to decrease its variance to the level of the 
LeGrad method. In this example, we have found empirically that this is achieved when LdGrad uses $S=10^4$ 
samples; see Figure \ref{fig:toyGaussian}(b). This shows that LeGrad is significantly better 
than LdGrad and the same conclusion is supported from the experiment in Section \ref{sec:logreg}
where we consider a Bayesian logistic regression model.  

Furthermore, observe that the fact that LeGrad has lower variance than ReGrad is in a good accordance with the 
theoretical results of Section \ref{sec:conneReparam}. Notice also that the variance of LeGrad is 
roughly one order of magnitude lower than the variance of ReGrad.  
  
Finally, to visualize the convergence of the different algorithms and their ability to maximize 
the lower bound in Figure  \ref{fig:toyGaussian}(c) we plot the stochastic value of the lower 
bound computed at each iteration by drawing a single sample from the variational distribution. Clearly, 
the stochastic value of the bound can allow us to quantify convergence and it could also be used as a diagnostic of  
high variance problems. From Figure \ref{fig:toyGaussian}(c) we can observe 
that LdGrad makes very slow progress in maximizing the bound while LeGrad and ReGrad converge 
rapidly. 

\subsection{Logistic regression \label{sec:logreg}}

In this section we compare the three approaches in a challenging 
binary classification problem using Bayesian logistic regression. We  
apply the stochastic variational algorithm in order to approximate the posterior 
over the regression parameters. Specifically, given a dataset 
$\dataset \equiv \{\bfz_j, y_j\}_{j=1}^m$, where $\bfz_m\in \mathbbm{R}^{n}$ 
is the input and $y_m \in \{-1,+1\}$ the class label, 
we model the joint distribution over the observed labels and the parameters $\bfw$ by  
$$
p(\bfy, \bfw) = \left( \prod_{m=1}^M \sigma(y_m \bfz_m^{\T} \bfw) \right) p(\bfw),
$$
where $\sigma(a)$ is the sigmoid function and $p(\bfw)$ denotes a zero-mean Gaussian prior 
on the weights $\bfw$. As in the previous section we will assume a variational 
Gaussian distribution defined as in eq.\ (\ref{eq:factorGauss}) with the only notational 
difference that the unknowns now are model parameters, denoted by $\bfw$, and not latent 
variables.  

For the above setting we considered a subset of the MNIST  dataset that includes all 
$12660$ training examples from the digit classes $2$ and $7$. We applied all three 
stochastic gradient methods using a setup analogous to the one used in the previous section. 
In particular, the LeGrad method uses again a Gaussian quadrature rule of $K=5$ grid points so that  the time 
complexity per iteration was $O(n K) = O(785 * 5)$ and where the number $785$ comes
from the dimensionality of the digit images plus the bias term. To match this with the time complexity 
of LdGrad, we will use a size of $S=3925$ samples in the Monte Carlo approximation in 
eq.\ (\ref{eq:logderivestimate}). 

\begin{figure*}[!htb]
\begin{center}
\begin{tabular}{ccc}
{\includegraphics[scale=0.3]
{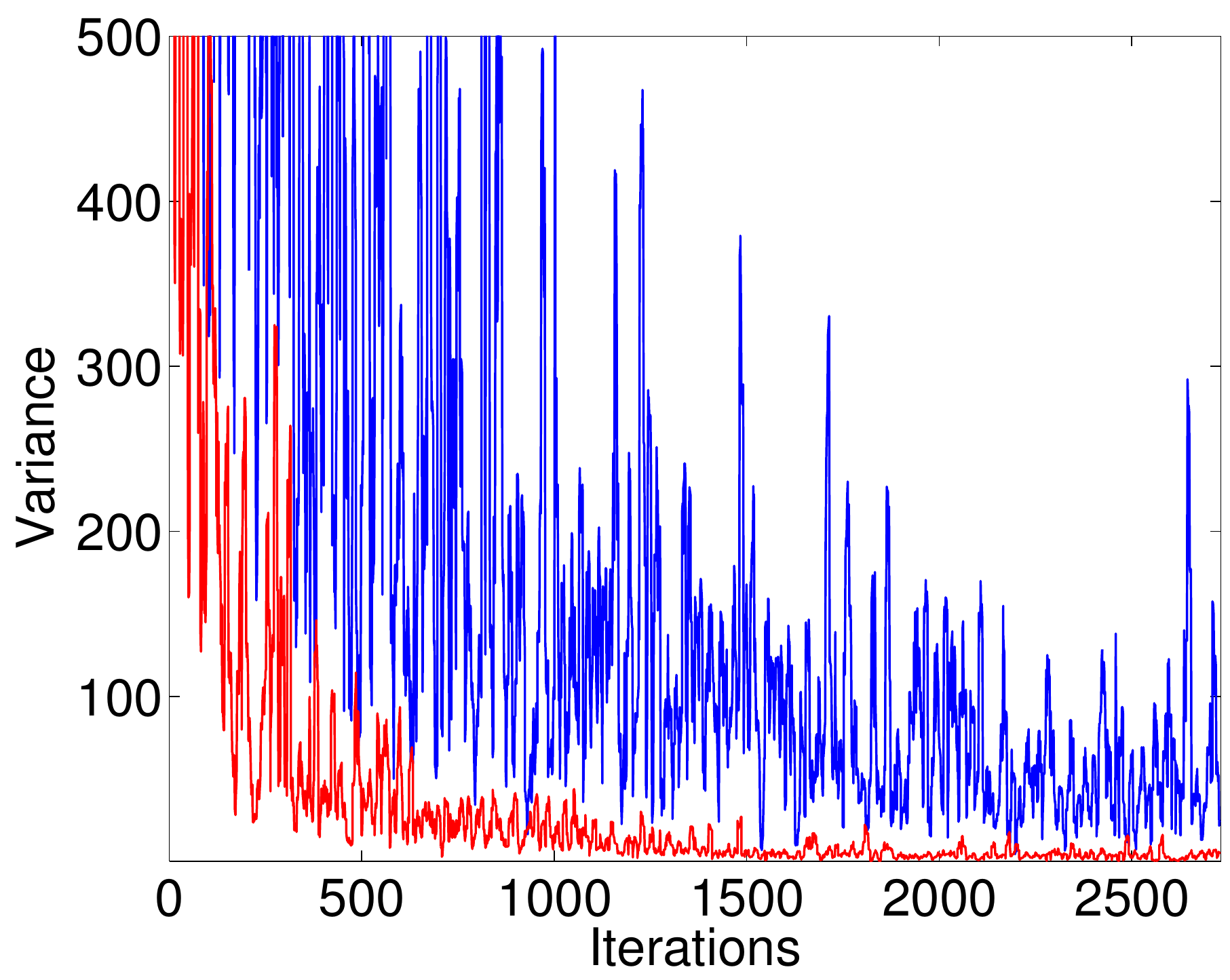}}  &
{\includegraphics[scale=0.3]  
{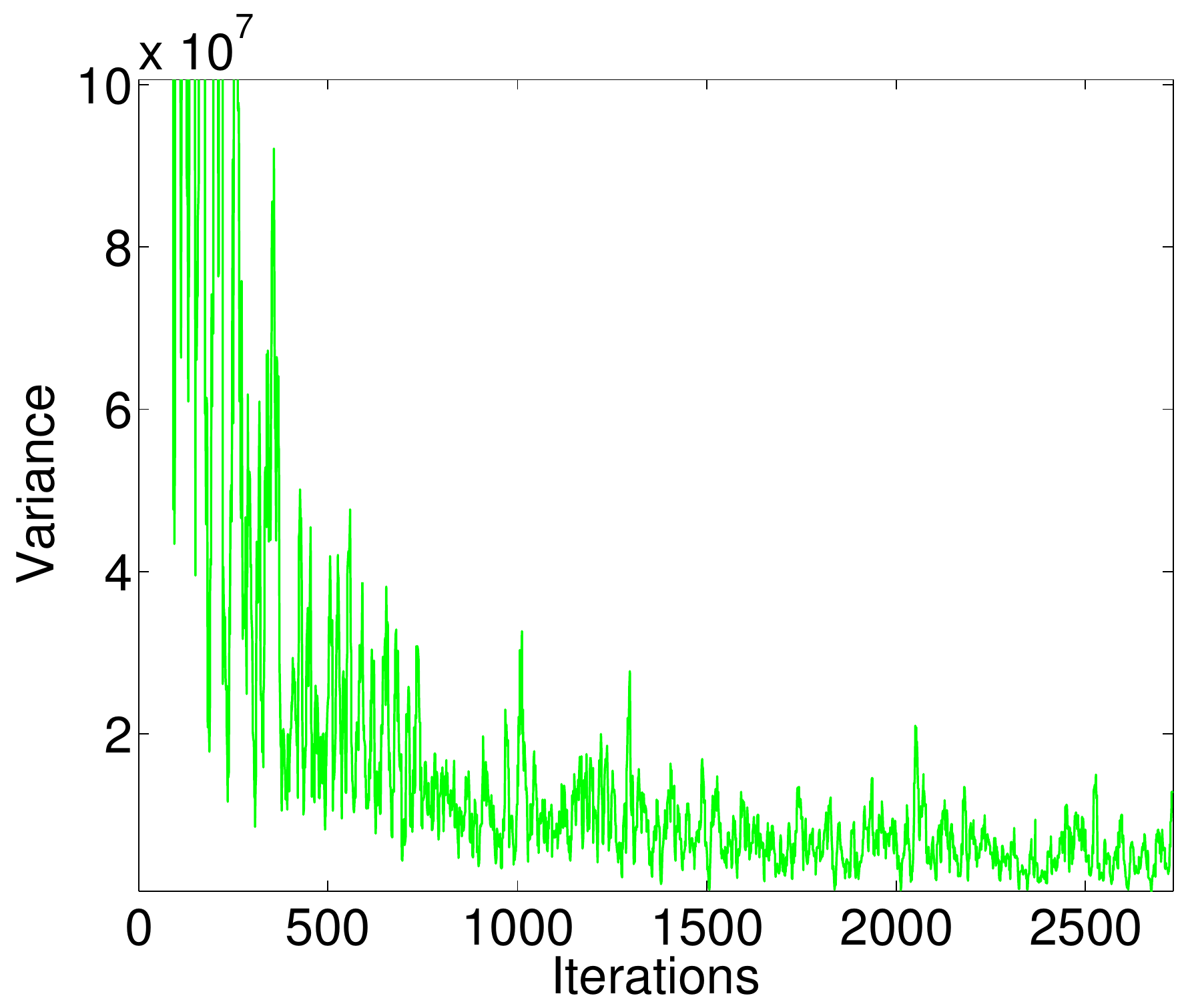}} &
{\includegraphics[scale=0.3]
{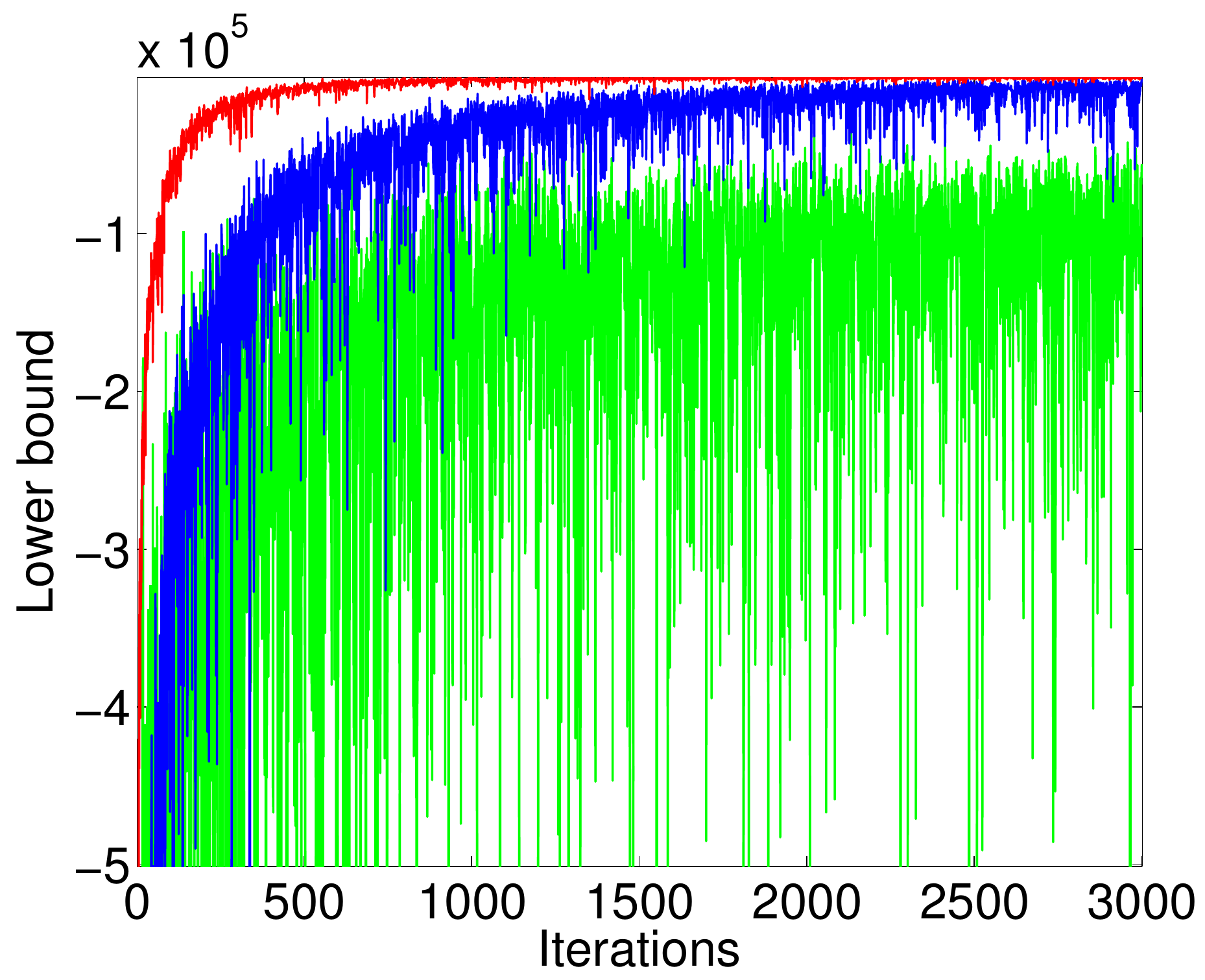}}
\end{tabular}
\caption{The panel in (a) shows the variance of the gradient for the variational parameter $\mu_1$ 
when using LeGrad (red line), ReGrad (blue line), while panel (b) shows the corresponding curve for LdGrad (green line). 
The number of samples used by LdGrad was $S=3925$. The panel in (c) shows the evolution of the stochastic values of the lower bound.} 
\label{fig:mnistLogreg}
\end{center}
\end{figure*}

Figure \ref{fig:mnistLogreg}(a) displays the variance of the stochastic gradient for the LeGrad method 
(green line) and the ReGrad method (blue line). As we can observe the local expectation method 
has roughly one order of magnitude lower variance than the reparametrization approach which is in a accordance with the 
results from the previous section (see Figure \ref{fig:toyGaussian}). In contrast to these two methods, which somehow have comparable variances,  
the LdGrad approach severely suffers from very high variance as shown in Figure \ref{fig:mnistLogreg}(b) where the displayed 
values are of the order of $10^7$. Despite the fact that $3925$ independent samples are used in the Monte Carlo approximation
still LdGrad cannot make good progress when maximizing the lower bound. To get a sensible performance with LdGrad 
we will need to increase considerably the sample-size which is computationally very expensive. 
Of course, control variates can be used to reduce variance to some extend, but it would have been much better if this 
problem was not present in the first place. LeGrad can be viewed as a certain variant of LdGrad 
but has the useful property that does not suffer from the high variance problem. 

Finally, Figure \ref{fig:mnistLogreg}(b) shows the evolution of the stochastic value of the lower bound
for all three methods. Here, we can observe that LeGrad has significant faster and much more stable convergence than 
the other two methods. Furthermore, unlike in the example from the previous section, in this real dataset the ReGrad 
method exhibits clearly much slower convergence that the LeGrad approach.

\begin{figure*}[!htb]
\vskip 0.2in
\begin{center}
\begin{tabular}{cc}
{\includegraphics[scale=0.45]
{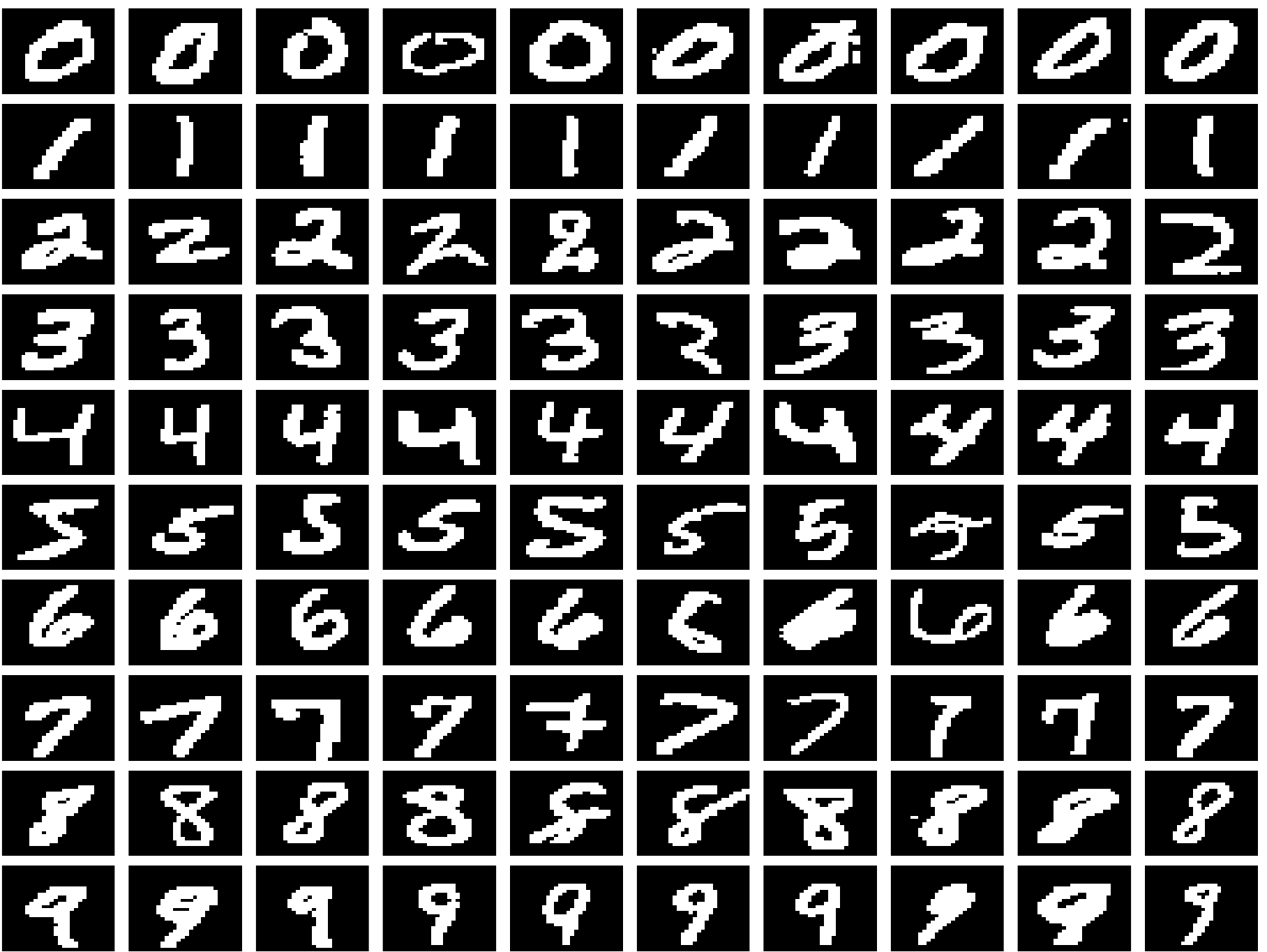}} &
{\includegraphics[scale=0.45]
{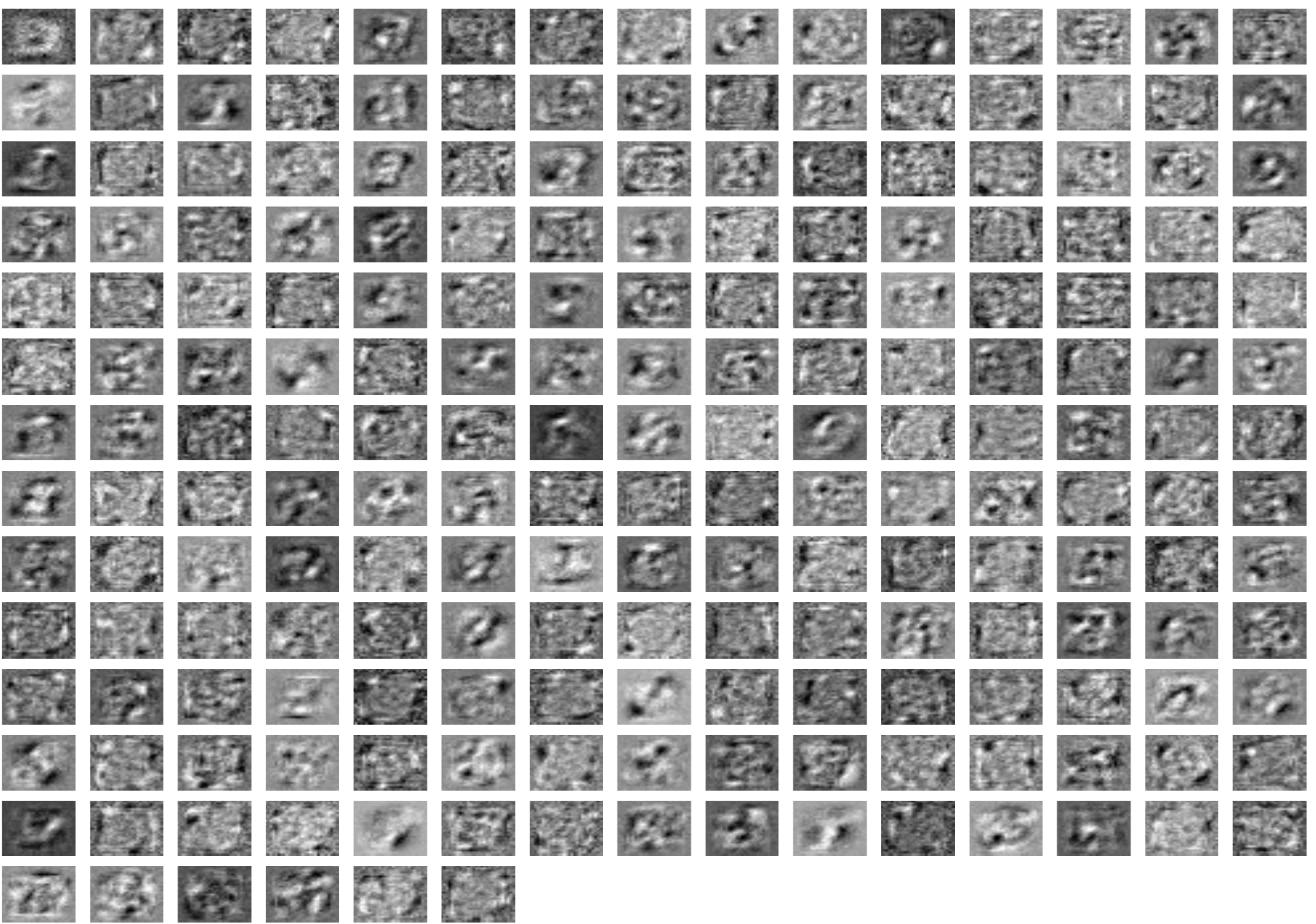}}  \\
{\includegraphics[scale=0.45]  
{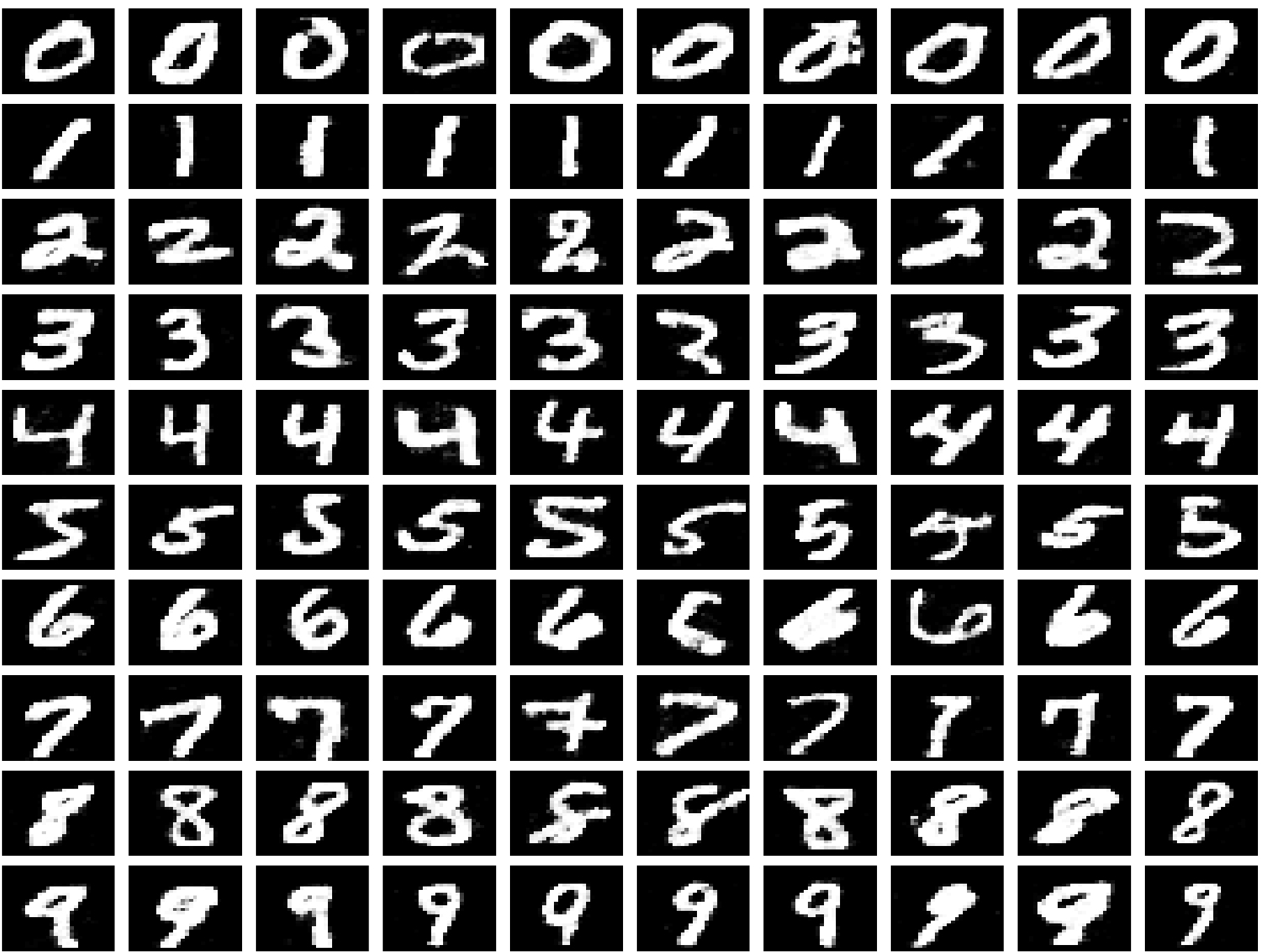}} & 
 {\includegraphics[scale=0.45]  
{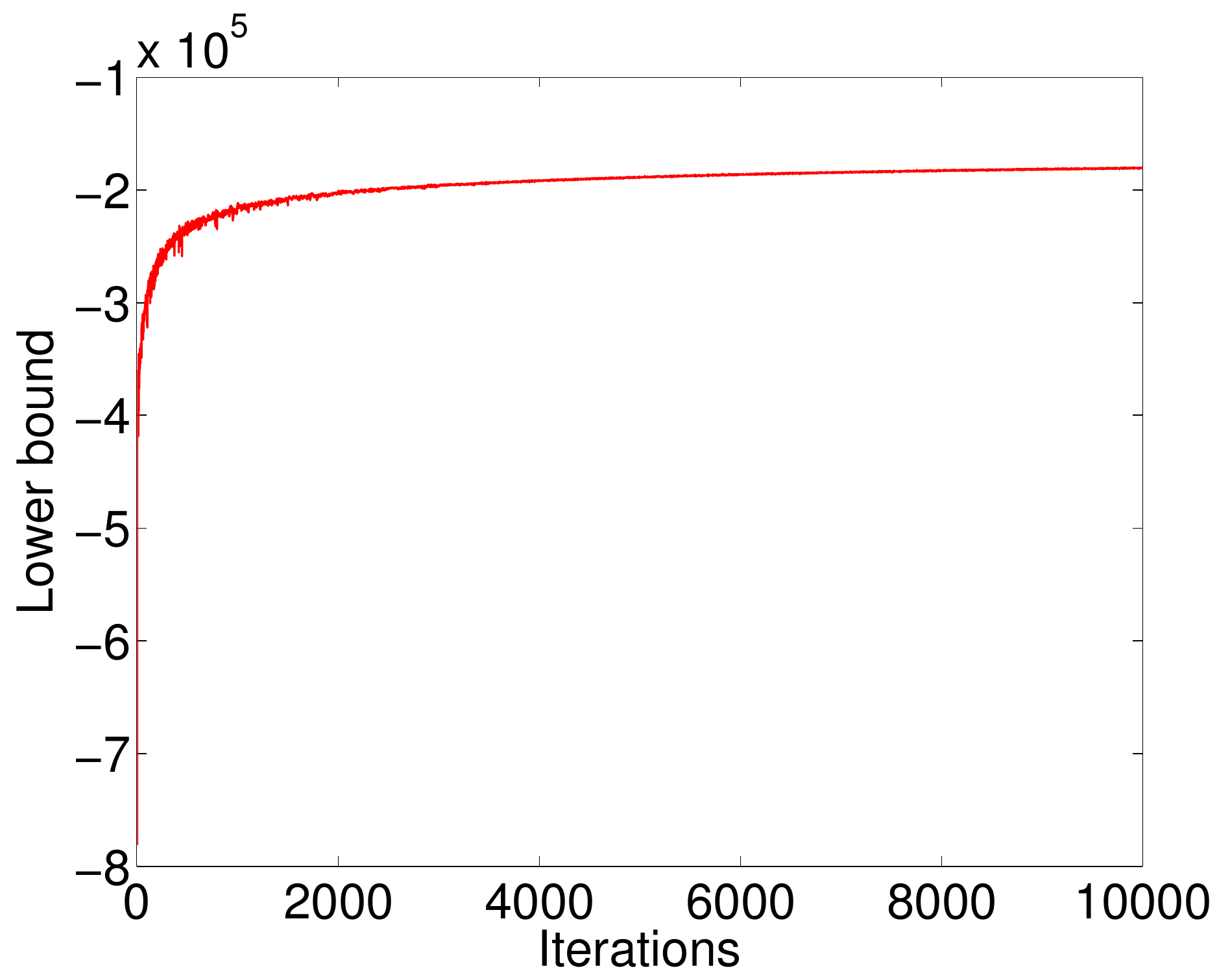}}
\end{tabular}
\caption{The first row shows $100$ training examples and the learned model parameters $W$ for all $K=200$ hidden variables. 
The second row shows the corresponding reconstructed data and the evolution of the stochastic value of the lower bound.} 
\label{fig:mnistSigmoid}
\end{center}
\vskip -0.2in
\end{figure*}


\subsection{Fitting one-layer sigmoid belief net with a recognition model \label{sec:sigmoidbnet}}

In the final example we consider a sigmoid belief network with a single hidden layer. 
More precisely, by assuming the observations are binary data vectors of the form $\bfy_i \in \{0,1\}^D$, 
such a sigmoid belief network assumes that each $\bfy_i$ is generated independently according to   
\begin{equation}
p(\bfy|W) = \sum_{\bfx}  \prod_{d=1}^D \left[ \sigma(\bfw_d^\T \bfx) \right]^{y_d} 
\left[1-\sigma(\bfw_d^\T \bfx) \right]^{1 -y_d}  p(\bfx), 
\label{eq:sigmoidbnJoint} 
\end{equation}
where $\bfx \in \{0,1\}^K$ is a vector of hidden variables while the prior $p(\bfx)$ is 
taken to be uniform. The matrix $W$ (that incorporates also a bias term) 
consists of the set of model parameters to be estimated by fitting the model to the data. In theory 
we could use the EM algorithm to learn the parameters $W$, however, such an approach is not 
feasible because at the E step we need to compute the posterior distribution $p(\bfx_i|\bfy_i,W)$
over each hidden variable which clearly is intractable since each $\bfx_i$ takes $2^K$ values. Therefore, 
we need to apply approximate inference and next we consider 
stochastic variational inference using the local expectation gradients algorithm. 

More precisely, by following recent trends in the literature for fitting this type of models 
\cite{MnihGregor2014, KingmaW13, stochbackpropDeepmind2014} we assume a variational
distribution parametrized by a ``reverse'' sigmoid network that predicts the latent vector $\bfx_i$ 
from the associated observation $\bfy_i$:
\begin{equation}
q_{V} (\bfx_i) = \prod_{k=1}^K \left[ \sigma(\bfv_k^\T \bfy_i) \right]^{x_{i k}} 
\left[1-\sigma(\bfv_k^\T \bfy_i) \right]^{1 - x_{i k}},  
\label{eq:sigmoidbnQ} 
\end{equation}
where $V$ is a matrix (that incorporates also a bias term) comprising the set of all variational 
parameters to be estimated. Often a variational distribution of the above form is referred 
to as the {\em recognition model} because allows to predict the activations of the 
hidden variables in unseen data $\bfy_*$ without needing to fit each time 
a new variational distribution. 

Based on the above model we considered a set of 
$1000$ binarized MNIST digits so that $100$ examples were chosen from 
each digit class. The application of stochastic variational inference 
is straightforward and boils down to constructing a separate lower bound for 
each pair $(\bfy_i,\bfx_i)$ having the form
\begin{align}
\mathcal{F}_i(V,W) & = \sum_{\bfx_i} q_V(\bfx_i) \log \prod_{d=1}^D
[\sigma(\bfw_d^\T \bfx_i)]^{y_{i d}} [1 - \sigma(\bfw_d^\T \bfx_i)]^{1-y_{i d}}  \nonumber \\
& - \sum_{k=1}^K \sigma(\bfv_k^\T \bfy_i) \log  \sigma(\bfv_k^\T \bfy_i) - \sum_{k=1}^K (1 - \sigma(\bfv_k^\T \bfy_i)) \log (1 - \sigma(\bfv_k^\T \bfy_i)). 
\end{align}
The total lower bound is then expressed as the sum of all data-specific bounds and it is 
maximized using stochastic updates to tune the forward model weights $W$ and the recognition weights 
$V$. Specifically, the update we used for $W$ was based on drawing a single sample from the full 
variational distribution: $q_{V}(\bfx_i)$, with $i=1,\ldots,n$. The update 
for the recognition weights was carried out by computing the stochastic gradients
according to the local expectation gradients so that for each $\bfv_k$ the 
estimate obtained the following form    
%
%
%
%
%
\begin{align}
\nabla_{\bfv_k} \mathcal{F} =  \sum_{i=1}^n \nabla_{\bfv_k} \mathcal{F}_i & = \sum_{i=1}^n \sigma_{ik} (1 - \sigma_{ik})  \left[ \sum_{d=1}^D \log \left( 
\frac{ 1 +  e^{- \widetilde{y}_{i d} \bfw_d^\T (\bfx_{i \setminus k}^{(t)}, x_{i k}=0) } }
{1 +  e^{- \widetilde{y}_{i d} \bfw_d^\T (\bfx_{i \setminus k}^{(t)}, x_{i k}=1) }} \right)   + \log \left( \frac{1 - \sigma_{i k}}{\sigma_{ik}} \right) \right] \bfy_i, 
\end{align}
where $\sigma_{i k} = \sigma(\bfv_k^\T \bfy_i)$ and $\widetilde{y}_{i d}$ is the $\{-1,1\}$ encoding of $y_{i d}$. Figure \ref{fig:mnistSigmoid} shows several plots that illustrate how the model fits the data  
and how the algorithm converges. More precisely,  we optimized the model assuming $K=200$ hidden variables in the hidden layer of the sigmoid network 
with the corresponding weights $W$ shown in Figure \ref{fig:mnistSigmoid}. 
Clearly, the model is able to provide very good reconstruction of the 
training data and exhibits also a fast and very stable convergence. 

Finally, for comparison reasons we tried also to optimize the variational lower bound using the LdGrad algorithm. However, this proved to be very problematic 
since the algorithm was unable to make good progress and has severe tendency to get stuck to local maxima possibly due to the high variance problems.    

\section{Discussion \label{sec:discussion}}

We have presented a stochastic variational inference algorithm which we call  
{\em local expectation gradients}. This algorithm provides a a general framework 
for estimating stochastic gradients that exploits the local independence structure 
of the variational distribution in order to efficiently optimize the variational parameters. 
We have shown that this algorithm does not suffer from high variance problems.   
Future work will concern with the further theoretical analysis of the properties of the algorithm as well 
as applications to hierarchical probabilistic models such as sigmoid belief networks with multiple layers.


\nocite{langley00}

\bibliography{svi}
\bibliographystyle{plain}

\end{document}